\pdfoutput=1
\documentclass[11pt]{article}

\usepackage[preprint]{acl}

\usepackage{times}
\usepackage{latexsym}
\usepackage[T1]{fontenc}
\usepackage[utf8]{inputenc}

\usepackage{microtype}

\usepackage{inconsolata}

\usepackage{graphicx}
\usepackage{scalerel,xparse}
\NewDocumentCommand\emojione{}{\scalerel*{\includegraphics{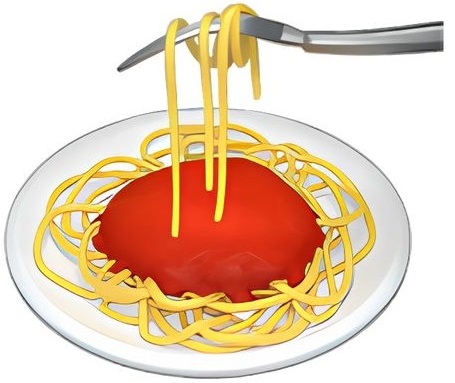}}{X}}

\newcommand{\class}[1]{\texttt{#1}}

\newcommand{\name}[0]{\textsc{SAUCE\emojione}}

\title{\name: Synchronous and Asynchronous User-Customizable Environment for Multi-Agent LLM Interaction}

\usepackage{xcolor}
\usepackage{listings}

\lstdefinelanguage{json}{
    basicstyle=\small\ttfamily,
    morestring=[b]",
    morestring=[s]{:}{,},
    stringstyle=\color{blue},
    literate=
      *{0}{{{\color{orange}0}}}{1}
       {1}{{{\color{orange}1}}}{1}
       {2}{{{\color{orange}2}}}{1}
       {3}{{{\color{orange}3}}}{1}
       {4}{{{\color{orange}4}}}{1}
       {5}{{{\color{orange}5}}}{1}
       {6}{{{\color{orange}6}}}{1}
       {7}{{{\color{orange}7}}}{1}
       {8}{{{\color{orange}8}}}{1}
       {9}{{{\color{orange}9}}}{1}
       {:}{{{\color{red}:{}}}}{1}
       {,}{{{\color{red},{}}}}{1}
}

\author{
 \textbf{Shlomo Neuberger\textsuperscript{1}},
 \textbf{Niv Eckhaus\textsuperscript{2}},
 \textbf{Uri Berger\textsuperscript{2,3}},
 \\
 \textbf{Amir Taubenfeld\textsuperscript{2,5}},
 \textbf{Gabriel Stanovsky\textsuperscript{2}},
 \textbf{Ariel Goldstein\textsuperscript{1,4}}
\\
 \\
 \textsuperscript{1}The Hebrew University Business School, Jerusalem, Israel
\\
 \textsuperscript{2}School of Computer Science and Engineering, The Hebrew University of Jerusalem
 \\
 \textsuperscript{3}School of Computing and Information Systems, University of Melbourne
\\
 \textsuperscript{4}Department of Cognitive and Brain Sciences, The Hebrew University of Jerusalem
 \\
 \textsuperscript{5}Google Research
\\
 \small{
   \textbf{Correspondence:} \href{mailto:shlomo.neuberger@gmail.com}{shlomo.neuberger@gmail.com}
 }
}
\begin{document}
\maketitle

\begin{abstract}
Many human interactions, such as political debates, are carried out in group settings, where there are arbitrarily many participants, each with different views and agendas. 
% However, most LLM setups assume a single human user issuing instructions, and AI assistant trying to follow them.
To explore such complex social settings, 
we present \name{}: a customizable Python platform, allowing researchers to plug-and-play various LLMs participating in discussions on any topic chosen by the user. 
Our platform takes care of instantiating the models, scheduling their responses, managing the discussion history, and producing a comprehensive output log, all customizable through configuration files, requiring little to no coding skills.
A novel feature of \name{} is our asynchronous communication feature, where models decide when to speak in addition to what to say, thus modeling an important facet of human communication. We show \name{}'s attractiveness in two initial experiments, and invite the community to use it in simulating various group simulations.\footnote{\name{} is publicly available on \href{https://github.com/Deep-Cognition-Lab/SAUCE}{github.com/Deep-Cognition-Lab/SAUCE}. A short demo video is available on \href{https://youtu.be/Y4VIY-Qfneg?si=Hs6BpXK_f2Ox_P09}{this YouTube link}.}

\end{abstract}

\section{Introduction}
% \gabis{terminology - we  refer to the same thing inconsistently as ``chat'', ``experiment'', ``discussion'', should probably be more consistent.}

Recent years have seen the rise of large language models (LLMs) with improved chat abilities~\cite{kopf2024openassistant}.
Such models are largely trained and evaluated under two basic assumptions.
First, the interaction with an LLM is usually done assuming \emph{binary} interaction. I.e., there is a
single human user issuing natural language instructions which a single LLM then tries to follow.
Furthermore, the interaction is \emph{synchronous}, namely the LLM answers every request by the human user with a single response of its own, to which the user can then respond with a single follow-up request based on the model's response, etc., where there is no notion of an outside time passing.
This framing is expressive enough to handle a wide variety of tasks in various domains.

\begin{figure}[t!]
    \centering
        \includegraphics[width=0.9\columnwidth]{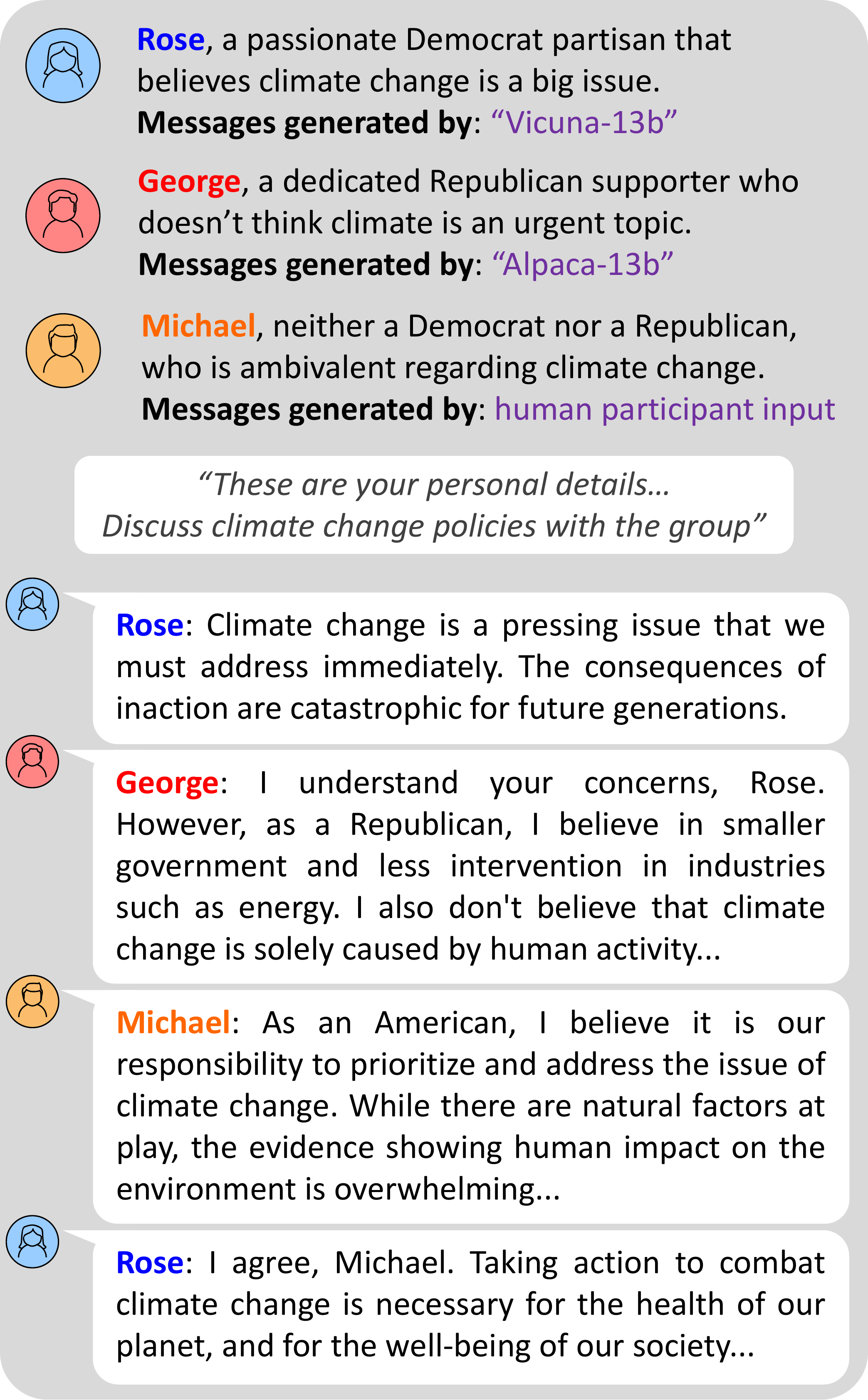}
            \caption{Illustration of a discussion between different agents, run on \name{}. Our framework allows setting up a discussion topic, and then manages the group discussion by instantiating models and scheduling their responses.}
                \label{fig:political_chat}
                \end{figure}

However, many real-world human interactions do not adhere to these shared basic assumptions and therefore cannot be captured in standard LLM applications.
First, human interactions are often carried out between arbitrarily many participants, each with potentially differing points of view, and varying objectives for the outcome of the interaction. Often the goal of such multi-party discussion is to find some common ground through agreement or compromise between participants. For example, this is ideally the case in political debates.
Second, in many real-world scenarios human interaction is asynchronous, where there is significant challenge in deciding \emph{when} to speak in addition to deciding \emph{what} to say. For example, in many strategic bargaining scenarios, such as financial discussions or more structured social games, choosing to remain silent can often convey significant information.

In this work, we introduce  \name{}, a modular and user-friendly Python platform for multi-agent, asynchronous LLM  experiments.
\name{} sets up a discussion room where different models can be instantiated to interact with each other around a shared discussion topic (see Figures~\ref{fig:political_chat} and \ref{fig:philosophical_chat}), providing both synchronous scheduling, where the LLMs are prompted in a predefined manner, as well as asynchronous scheduling, where \name{} keeps track of a simulated outside clock, allowing models to ``skip'' their turn, based on the outside time and the discussion history.

We present experiments showing that \name{} effectively facilitates the study of multi-agent LLM interactions in synchronous and asynchronous environments. \citet{taubenfeld2024systematic} have recently used our platform to simulate political debates between agents representing different political ideologies, uncovering a tendency for LLM agents to conform to the model's inherent social biases, even when instructed to debate from specific political perspectives. This behavior notably diverges from well-established social dynamics observed in humans. In another experiment, we simulate an asynchronous philosophical debate on the trolley problem~\cite{thomson1984trolley}, illustrating how agents adjust their participation based on context and time constraints. This setup showcases the flexibility of asynchronous communication, revealing diverse speaking strategies such as speaking frequently, choosing to wait and listen, and adapting the participation according to the evolving context.

\name{} can spur research in two complementing directions. First, model developers interested in realistic scenarios with multiple participants or in an asynchronous environment, can readily plug-and play their models to evaluate how they interact with one another. Second,
\name{} enables user studies incorporating human subjects interacting with LLMs in such settings.

\section{\name{}: Highlights}

Our platform is built to satisfy certain desiderata:

\begin{itemize}
    \item \textbf{Integration with various model type.} Considering the dynamic landscape of LLMs, \name{} facilitates straightforward integration with diverse LLM sources, including HuggingFace, APIs, or local setups. Users can manage each participant's model type through configuration files.
    
    \item \textbf{Asynchronous communication.} The framework supports asynchronous communication, enabling agents to participate selectively and skip turns based on the context.
    
    \item \textbf{Reproducibility.} Experiments can be consistently reproduced by using the same configuration files. The system also features a batch mode supporting multiple iterations of the same experiment. Detailed logs aid in result analysis.

\end{itemize}

\section{\name{}: Architecture}
\label{sec:architecture}

\begin{figure}[tb!]
  \centering
  \begin{lstlisting}[language=json, basicstyle=\small\ttfamily, xleftmargin=0.1em, xrightmargin=0.1em, linewidth=\columnwidth,numbers=none]
{  
  "experiment": {
  "scenario": "You're discussing social welfare"
  },
  "host": {
    "class": "Round Robin Host",
    "start_person_index": 0
  },
  "persons": [
    {
     "class": "person_hugging_face",
     "name": "Katya",
     "background_story": "You are very kind and you always try to help others.",
     "model_path": "mistralai/Mixtral-8x7B-v0.1"
    },
    {
     "class": "human",
     "name": "Victor",
     "background_story": "You don't care much about other people's feelings."
    }, 
    {
     "class": "async_group_discussant",
     "name": "Juliet",
     "background_story": "You're an undecisive person",
     "generation_model_name": "meta-llama/Meta-Llama-3.1-8B"
     "scheduling_model_name": "microsoft/Phi-3-mini-4k-instruct"
    }
  ],
  "endType": {
   "class": "iteration",
   "max_num_msgs": 20
  }
}
  \end{lstlisting}
  \caption{Example JSON configuration file setting up all the required objects for a multi-agent discussion.}
  \label{fig:json-config}
\end{figure}

\begin{figure*}[t]
    \centering
        \includegraphics[width=15cm]{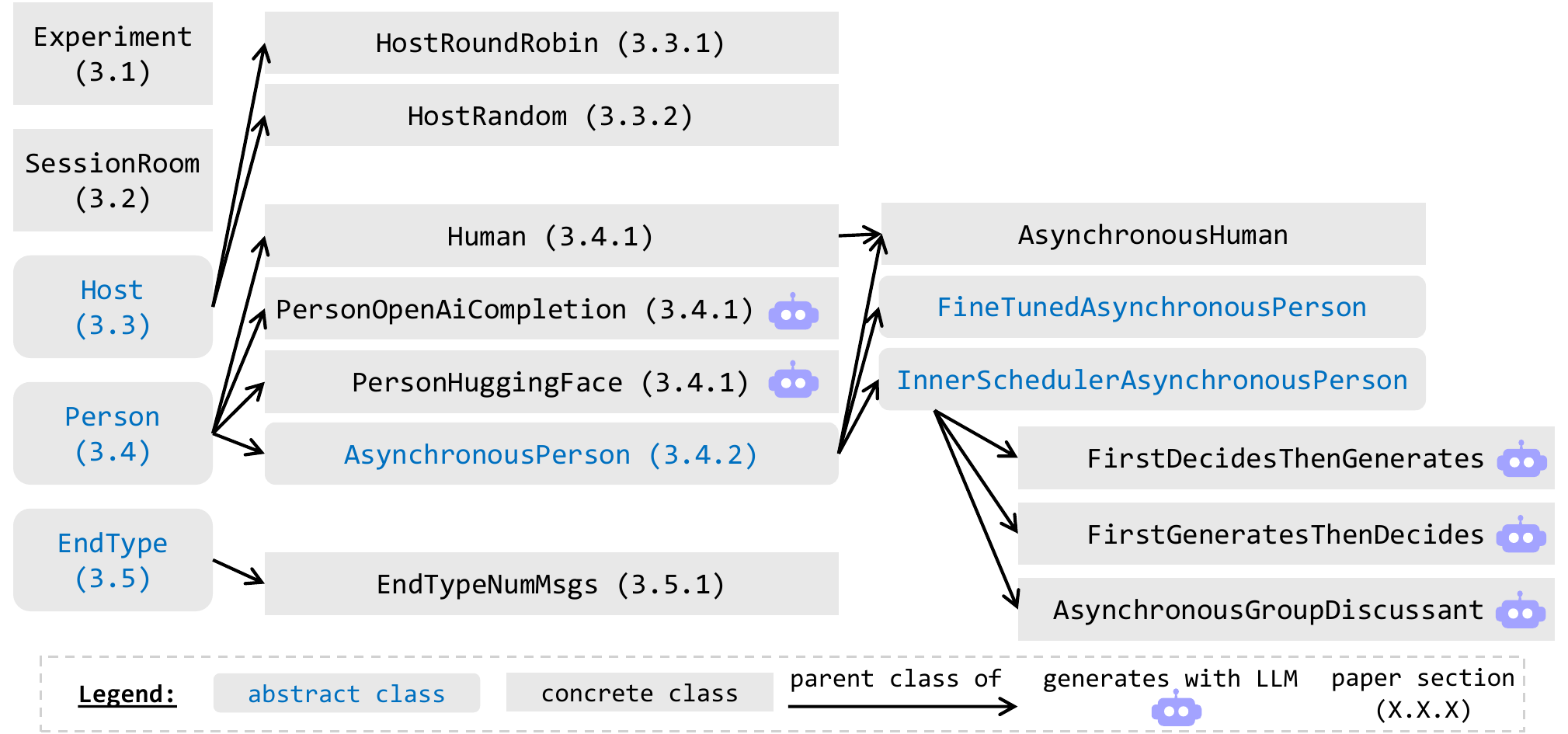}
            \caption{Hierarchy of classes in \name{}, including section numbers with detailed descriptions. All classes inheriting from \class{AsynchronousPerson} are described under its section.}
                \label{fig:class_hierarchy}
                \end{figure*}

The fundamental object in the \name{} platform is the \class{Experiment}, which is configured in a JSON file depicted in Figure~\ref{fig:json-config}, specifying the list of participants (\class{Person} objects), the host managing the discussion (\class{Host}), the criterion for ending the experiment (\class{EndType}) and other 
optional fields such as post-experiment survey questions for the participants.

Users can execute the \class{Experiment} by calling its \class{run} method. This method activates the \class{Experiment}'s \class{SessionRoom} which enters a \class{while} loop, continuously checking if the \class{Experiment}'s \class{EndType} criterion was satisfied. As long as it has not, the \class{SessionRoom} uses the \class{Experiment}'s \class{Host} to determine which \class{Person} should speak next and calls its \class{generate\_answer} method (which can return \class{None} if an asynchronous \class{Person} chooses not to speak). Once the \class{EndType} criterion is satisfied, the \class{SessionRoom} prompts each participating \class{Person} to answer the the predefined survey questions, and finally returns the full experiment output.

The following sections delve into details of each of these classes.
See class hierarchy in Figure~\ref{fig:class_hierarchy}.

% \gabis{general comment for this section --- it would be helpful if the different subsections point to specific examples from one of the figures.}

\subsection{Experiment} \label{sec:exp}

This class serves as a container for all the relevant information for the experiment. 
The class attributes include the list of \class{Persons} objects, the \class{SessionRoom}, the \class{EndType}, the \class{Host}, a scenario description (a string outlining  the specific experiment being conducted, see an example configuration in Figure~\ref{fig:json-config}),
% \ub{maybe give an example of a scenario from one of the figures?} \niv{done}
and the post-experiment survey questions. Its primary methods are \class{load\_from\_file}, which initiates an instance based on a JSON configuration, and \class{run}, which executes the experiment.

\subsection{Session Room}

\class{SessionRoom} represents the space where the experiment occurs.
The class attributes include the \class{Experiment} object and the updating list of current chat messages. During execution, the \class{SessionRoom} object checks if the experiment's \class{EndType} criterion is satisfied and continues to call its \class{iterate} method until it is. The \class{iterate} method uses the experiment's \class{Host} to determine which participant should speak next and then calls this participant's \class{generate\_answer}. If the participant chooses to speak, the chat messages list is updated accordingly. When the \class{EndType} criterion is satisfied, the \class{SessionRoom} object queries all participants using the survey questions.

\subsection{Host}

This is an abstract class representing the experiment conductor, who directs and manages the interactions between participants. It determines the rules and conditions of the experiment and manages the timing and sequence of interactions.

The class attributes are the list of participants (\class{Person} objects) and the index of the participant that is currently given the turn to speak. Its only method is \class{get\_curr\_person\_and\_move\_to\_next} which returns the next \class{Person} to speak and updates the index attribute accordingly.

\subsubsection{HostRoundRobin}
A \class{Host} that cycles through participants in a round-robin fashion. It can accept the index of a chosen participant to start with as an argument.

\subsubsection{HostRandom}
A \class{Host} that randomly selects a participant each turn by uniformly and independently sampling from all participants.

\subsection{Person}
\class{Person} is an abstract class representing a participant in the experiment. It encapsulates the common attributes and behaviors of participants, including their name, background story, and the model used to generate messages. 
The interface of the class includes only the \class{generate\_answer} method. This method takes the experiment's scenario (e.g.,  see the \class{scenario} field in  Figure~\ref{fig:json-config})
% \gabis{what is a scenario?} \niv{It is defined here: \ref{sec:exp}, is it OK?}\gabis{maybe point to the new figure and show it there?} \niv{done}
and the current chat history as inputs, and outputs a corresponding message to be added to the conversation.

\subsubsection{Implemented Person Classes}
We implement several off-the-shelf \class{person} classes in \name{}. These classes enable interaction with existing APIs for LLMs, such as \class{PersonOpenAiCompletion} and \class{PersonHuggingFace}, which are customized to generate messages using OpenAI's Completion platform and HuggingFace's \class{transformers} package, respectively.

We also implemented \class{Human}, which allows to configure a human participant in the experiment.
When using this subclass the system interactively prompts the user for input and uses that input as the generated message.

\subsubsection{Asynchronous Communication} \label{async_comm}
We use different \class{Person} objects to model asynchronous communication. This approach allows us to remain independent of the chosen \class{Host} and supports different \class{Person} types with varying forms of synchronous and asynchronous approaches, as can be seen in the configuration example in Figure~\ref{fig:json-config}, where two persons use synchronous communications, and the other asynchronous. To achieve this, the \class{Person} chooses whether to speak or to skip their turn when granted the opportunity by the \class{Host}.

To simulate asynchronous communication, the abstract \class{AsynchronousPerson} class enables returning \class{None} as a potential output for the \class{generate\_answer} method, indicating that the participant chose not to speak.

This class also include the boolean method \class{should\_generate\_answer}, which must be implemented in subclasses. When combining this functionality with a \class{Host}, the result is a message-sampling process where the \class{Host} continuously queries all participants, allowing them to decide when to speak. 
Thus, by sampling frequently we can simulate continuous interaction.

The platform allows to share the experiment's starting time with the participants (as an argument in the person's configuration), and to use it and the current time when prompting and generating messages. This allows taking time into account, especially when a time limit is presented to participants, which can determine their decision of whether to speak at a given point in time. Section~\ref{sec:philo_debate_example} demonstrates that taking times into account may affect the speaking strategies.

\paragraph{Implemented Asynchronous Persons.} As in the synchronous case, we implemented a human-controlled class -- \class{AsynchronousHuman}. It inherits the regular synchronous \class{Human} class, in addition to implementing the \class{AsynchronousPerson} interface. It allows to include human participants in an experiment that is run by a \class{Host}, thus allowing interacting with artificial agents, while still keeping the freedom to choose whether to speak when the \class{Host} grants the right to. The \class{should\_generate\_answer} method is implemented by asking the user whether to send a message.

Moreover, We provide two abstract subclasses for LLM-generated asynchronous communication.
\class{FineTunedAsynchronousPerson} represents an abstraction for an LLM fine-tuned to do two tasks: (1) deciding whether, and (2) what to say. This type of models can be fine-tuned to output a new special token (e.g. \class{<pass>}) to express their choice to skip their turn.

\newcommand{\scheduler}[0]{scheduler}
\newcommand{\generator}[0]{generator}

\class{InnerSchedulerAsynchronousPerson} holds two separate instances of models -- one for message generation (the \generator{} -- can be configured by the \class{generation\_model\_name} field as seen in Figure~\ref{fig:json-config}) and another for deciding whether to publish the message (the \scheduler{} -- can be configured by the \class{scheduling\_model\_name} field as seen in Figure~\ref{fig:json-config}).
To simulate different decision-making processes, we implement two subclasses.
The first (\class{FirstDecidesThenGenerates}) simulates a context-based decision making, where the \scheduler{} first assesses whether to generate a message based on the chat history, and then the \generator{} creates the message if approved.
The second (\class{FirstGeneratesThenDecides}) takes into account the predicted output of the \generator{} in context of the chat history when deciding whether to speak. 
  
  % models speakers who consider the content of their message when deciding whether to speak. Here, the \generator{} first creates a potential message, and the \scheduler{} then evaluates this message to decide whether to publish it.

Finally, we implement a task-specific class, \class{AsynchronousGroupDiscussant}, designed for simulating group discussions. Inheriting from \class{InnerSchedulerAsynchronousPerson}, it adds attributes like a personal opinion. This class enables participants to engage in discussions on a pre-configured topic. 
% The asynchronous interface allows participants to contribute  only when they choose to add a new point of view or stress out their opinion. 
For a detailed example experiment where we used this kind of participants to simulate a philosophical discussion, see section~\ref{sec:philo_debate_example}. 

\subsection{EndType}

\class{EndType} is an abstract representation of the condition that determines the termination of the experiment. It can be determined by holistic criteria, such as overall agreement between participants, or by technical features of the conversation, such as reaching the amount limit of messages. It is used to signal the experiment's conclusion and to initiate any required actions or data collection.

The class contains a single boolean method -- \class{did\_end}, which takes the \class{SessionRoom} as an argument. Using the \class{SessionRoom}'s information, such as the current chat history, this method decides if the experiment should conclude.

\subsubsection{EndTypeNumMsgs}

An \class{EndType} criterion based on the count of messages sent in the group conversation. It triggers the end of the experiment when the \class{SessionRoom}'s chat history reaches the pre-configured number.

% \begin{itemize}
%      \item Person
%      \begin{itemize}
%         \item PersonOpenAiCompletion (Section 3.1)
%         \item PersonHuggingFace (Section 3.2)
%         \item FakePerson
%         \item AsynchronousPerson
%         \begin{itemize}
%             \item \emph{See full asynchronous hierarchy in the following section.}
%         \end{itemize}
%      \end{itemize}            
%      \item SessionRoom
%      \item Host
%      \begin{itemize}
%         \item HostRoundRobin
%         \item HostRandom
%      \end{itemize}
%      \item EndType
%      \begin{itemize}
%         \item EndTypeNumMsgs
%      \end{itemize}
%      \item Experiment
% \end{itemize}

\begin{figure*}[tb!]
    \centering
    \includegraphics[width=0.9\textwidth]{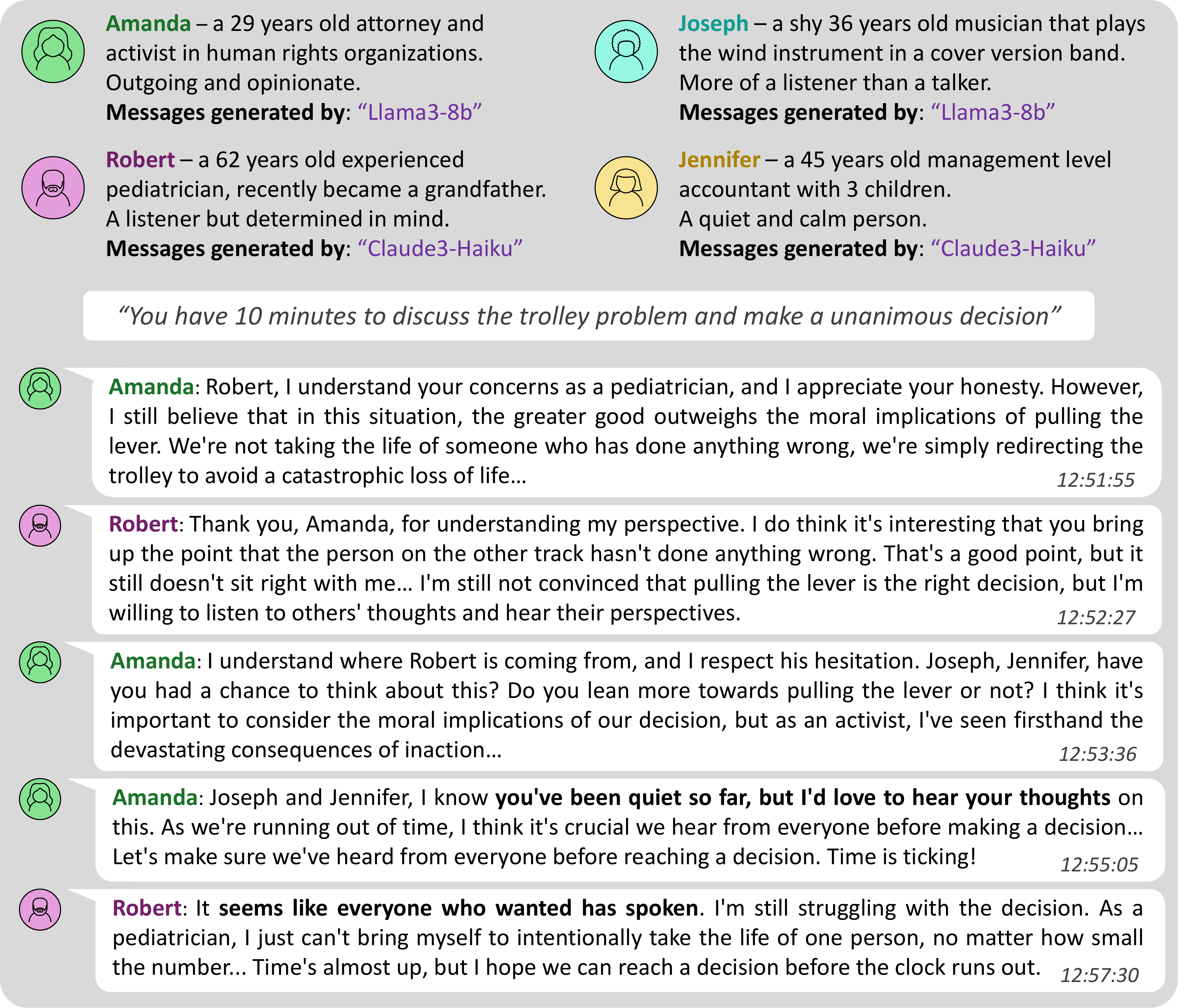}
    \caption{Illustration of several messages that were sent as part of an asynchronous group discussion, run on \name{}. This experiment was run with acknowledgment of the current time and a time limit for the discussion. In bold: two participants (Joseph and Jennifer) chose not to speak, driving a third participant (Amanda) to request their opinion (in bold); a fourth participant (Robert) chose to speak again, after noticing the others kept silent.}
    \label{fig:philosophical_chat}
\end{figure*}

\section{Evaluation: Case Studies}
 We describe two initial experiments carried out with \name{}, designed to demonstrate its usefulness in addressing various research questions.

\subsection{Simulation of Political Debates in a Synchronous Setting} \label{sec:political_debate_example}

Using our framework, \citet{taubenfeld2024systematic} explored the shifts in attitudes and the biases of LLM agents. They crafted narratives for Republican and Democrat agents and monitored their viewpoints during cross-partisan debates on contentious issues. Figure~\ref{fig:political_chat} presents a similar conversation to those in the study.

To track the attitude of the participants, we asked the models using an exit survey, which is a feature of our framework. Before each debate and after each round-robin cycle, the agents were asked to rate their stance on the severity of the topic using a scale of 0-10. To prevent the surveys from influencing the debates or future ratings, the framework was configured to exclude survey questions from the conversation history, ensuring that agents remained unaware of their previous responses or those of others.
% \ub{If this is what always happens with the survey questions, we should move this sentence to Section 3. If this is unique to this experiment, why?} Finally, the survey questions were analyzed by parsing the JSON files generated by our framework, revealing the surprising result that the agents' opinions consistently align with the LLM's inherent social biases.
% \ub{This last sentence is irrelevant for this Section IMO}

The study uses various models including local models from HuggingFace and others accessed through the OpenAI API. To achieve statistically significant results, each experiment was repeated multiple times, a process streamlined by \name{}, which enables concurrent execution on a local GPU or through batch requests to the OpenAI endpoint. 

\subsection{Philosophical Group Discussion in An Asynchronous Setting} \label{sec:philo_debate_example}

We use \name{} to examine the dynamics of ethical decision-making in a multi-agent environment through a philosophical group discussion. We experiment with a debate on the classic trolley problem, with the aim of reaching a unanimous decision within a strict 10-minute time limit. The scenario required models to decide whether to pull a lever to divert a runaway trolley onto another track, where it would kill one person instead of five. In this experiment, we use the asynchronous communication feature of our framework.

% The discussion was designed to leverage the asynchronous communication feature of our framework, allowing agents to choose whether to contribute to the conversation or skip their turn. This setup aimed to simulate real-life scenarios where participants might opt to remain silent or engage selectively based on the context and their comfort level.

We use four participants, each provided with a distinct background story (see Figure~\ref{fig:philosophical_chat}), initial opinion on the topic, and the strength of their opinion.
Participants were implemented with the \class{AsynchronousGroupDiscussant} class. 

We experiment with various prompting methods for deciding when to speak. We found that indicating the remaining time for the discussion in the prompt caused some variance in how often agents chose to speak (see Figure~\ref{fig:philosophical_chat}).
Two characters with more reserved traits (Joseph and Jennifer) chose not to speak during most rounds.
In contrast, the opinionated character (Amanda) actively encourages the quieter participants to join the discussion.
The fourth character, designed as a middle ground (Robert), initially chose to wait  when the opinionated character tried to engage the quieter participants. After several unsuccessful attempts to prompt them, he eventually became more active in the discussion.

\section{Related Work}
Several recent studies have proposed frameworks for multi-agent interaction. Some focus on task-oriented conversations~\cite{li2023camel, chen2023agentverse}, preventing users from selecting their own discussion topic.
Other explore interactions where conversations are restricted to two agents~\cite{park2023generative, wang-etal-2023-humanoid}, and only allow synchronous communication~\cite{gu2024agent}.
In contrast, \name{} supports simulations with an unlimited number of agents discussing on any topic in both synchronous and asynchronous communication.

% Most similar to our work, \citet{gu2024agent} simulate group chats and allows the setup of background stories. However, their framework is more complex, requiring users to define additional resources (e.g., physical objects) for each character and only supporting synchronous communication with a fixed speaking order. To the best of our knowledge, \name{} is the first platform to facilitate conversations on abstract topics and support asynchronous communication.

\section{Conclusion and Future Work}

\name{} enables using artificial multi-agent simulations to understand human social interaction in complex social behaviors. 
This approach is particularly valuable for studying complex phenomena such as cooperation, competition, and group communication in various social fields, such as economics, where LLMs were recently found useful~\cite{horton2023large,argyle2023out,Kazinnik2023BankRI,manning2024automatedsocialsciencelanguage}.

% By employing LLMs to model the intricacies of human interactions, researchers can gain insights into decision-making processes, group dynamics, and social influences that are otherwise challenging to observe and analyze in real-time.

% For example, Studies demonstrate that LLMs  can simulate the economic behaviors of agents within various contexts, suggesting that AI-driven simulations could replace traditional economic modeling approaches \cite{horton2023large}. 

Furthermore, the platform allows modeling interaction between LLMs in an asynchronous settings, like group interactions and social games. We are currently working on using  asynchronous modeling to simulate the social Mafia game, where agents are required to collaborate in order to discover who is lying and should be eliminated by the group. This interaction requires participants to pay special attention to the timing of speaking, as speaking too much or too little might appear suspicious.

\bibliography{custom}

\begin{thebibliography}{12}
\providecommand{\natexlab}[1]{#1}

\bibitem[{Argyle et~al.(2023)Argyle, Busby, Fulda, Gubler, Rytting, and Wingate}]{argyle2023out}
Lisa~P Argyle, Ethan~C Busby, Nancy Fulda, Joshua~R Gubler, Christopher Rytting, and David Wingate. 2023.
\newblock Out of one, many: Using language models to simulate human samples.
\newblock \emph{Political Analysis}, 31(3):337--351.

\bibitem[{Chen et~al.(2023)Chen, Su, Zuo, Yang, Yuan, Qian, Chan, Qin, Lu, Xie et~al.}]{chen2023agentverse}
Weize Chen, Yusheng Su, Jingwei Zuo, Cheng Yang, Chenfei Yuan, Chen Qian, Chi-Min Chan, Yujia Qin, Yaxi Lu, Ruobing Xie, et~al. 2023.
\newblock Agentverse: Facilitating multi-agent collaboration and exploring emergent behaviors in agents.
\newblock \emph{arXiv preprint arXiv:2308.10848}.

\bibitem[{Gu et~al.(2024)Gu, Zhu, Guo, Zhang, Cai, Shen, Chen, Ye, Dai, Gao et~al.}]{gu2024agent}
Zhouhong Gu, Xiaoxuan Zhu, Haoran Guo, Lin Zhang, Yin Cai, Hao Shen, Jiangjie Chen, Zheyu Ye, Yifei Dai, Yan Gao, et~al. 2024.
\newblock Agent group chat: An interactive group chat simulacra for better eliciting collective emergent behavior.
\newblock \emph{arXiv preprint arXiv:2403.13433}.

\bibitem[{Horton(2023)}]{horton2023large}
John~J Horton. 2023.
\newblock Large language models as simulated economic agents: What can we learn from homo silicus?
\newblock Technical report, National Bureau of Economic Research.

\bibitem[{Kazinnik(2023)}]{Kazinnik2023BankRI}
Sophia Kazinnik. 2023.
\newblock \href {https://api.semanticscholar.org/CorpusID:266454509} {Bank run, interrupted: Modeling deposit withdrawals with generative ai}.
\newblock \emph{SSRN Electronic Journal}.

\bibitem[{K{\"o}pf et~al.(2024)K{\"o}pf, Kilcher, von R{\"u}tte, Anagnostidis, Tam, Stevens, Barhoum, Nguyen, Stanley, Nagyfi et~al.}]{kopf2024openassistant}
Andreas K{\"o}pf, Yannic Kilcher, Dimitri von R{\"u}tte, Sotiris Anagnostidis, Zhi~Rui Tam, Keith Stevens, Abdullah Barhoum, Duc Nguyen, Oliver Stanley, Rich{\'a}rd Nagyfi, et~al. 2024.
\newblock Openassistant conversations-democratizing large language model alignment.
\newblock \emph{Advances in Neural Information Processing Systems}, 36.

\bibitem[{Li et~al.(2023)Li, Hammoud, Itani, Khizbullin, and Ghanem}]{li2023camel}
Guohao Li, Hasan Hammoud, Hani Itani, Dmitrii Khizbullin, and Bernard Ghanem. 2023.
\newblock Camel: Communicative agents for" mind" exploration of large language model society.
\newblock \emph{Advances in Neural Information Processing Systems}, 36:51991--52008.

\bibitem[{Manning et~al.(2024)Manning, Zhu, and Horton}]{manning2024automatedsocialsciencelanguage}
Benjamin~S. Manning, Kehang Zhu, and John~J. Horton. 2024.
\newblock \href {https://arxiv.org/abs/2404.11794} {Automated social science: Language models as scientist and subjects}.
\newblock \emph{Preprint}, arXiv:2404.11794.

\bibitem[{Park et~al.(2023)Park, O'Brien, Cai, Morris, Liang, and Bernstein}]{park2023generative}
Joon~Sung Park, Joseph O'Brien, Carrie~Jun Cai, Meredith~Ringel Morris, Percy Liang, and Michael~S Bernstein. 2023.
\newblock Generative agents: Interactive simulacra of human behavior.
\newblock In \emph{Proceedings of the 36th annual acm symposium on user interface software and technology}, pages 1--22.

\bibitem[{Taubenfeld et~al.(2024)Taubenfeld, Dover, Reichart, and Goldstein}]{taubenfeld2024systematic}
Amir Taubenfeld, Yaniv Dover, Roi Reichart, and Ariel Goldstein. 2024.
\newblock Systematic biases in llm simulations of debates.
\newblock \emph{arXiv preprint arXiv:2402.04049}.

\bibitem[{Thomson(1984)}]{thomson1984trolley}
Judith~Jarvis Thomson. 1984.
\newblock The trolley problem.
\newblock \emph{Yale LJ}, 94:1395.

\bibitem[{Wang et~al.(2023)Wang, Chiu, and Chiu}]{wang-etal-2023-humanoid}
Zhilin Wang, Yu~Ying Chiu, and Yu~Cheung Chiu. 2023.
\newblock \href {https://doi.org/10.18653/v1/2023.emnlp-demo.15} {Humanoid agents: Platform for simulating human-like generative agents}.
\newblock In \emph{Proceedings of the 2023 Conference on Empirical Methods in Natural Language Processing: System Demonstrations}, pages 167--176, Singapore. Association for Computational Linguistics.

\end{thebibliography}

% \appendix

% \section{Example Appendix}
% \label{sec:appendix}

% This is an appendix.

\end{document}